\pdfoutput=1

\documentclass[11pt]{article}

\usepackage{emnlp2021}

\usepackage{ctable}
\usepackage{bm}
\usepackage{graphicx}
\usepackage{amssymb}
\usepackage[ruled,vlined]{algorithm2e}
\SetAlFnt{\small}
\usepackage{amsthm}
\newtheorem{theorem}{Theorem}[section]
\newtheorem{lemma}[theorem]{Lemma}
\usepackage{multirow}

\usepackage{times}
\usepackage{latexsym}

\usepackage[T1]{fontenc}

\usepackage[utf8]{inputenc}

\usepackage{microtype}

\title{Conical Classification For Computationally Efficient One-Class Topic Determination}

\author{Sameer Khanna \\
  Department of Computer Science, Stanford University, USA \\
  Department of Research and Development, Fortinet, USA \\
  \texttt{sameerk@stanford.edu} \\
}

\begin{document}
\maketitle
\begin{abstract}
As the Internet grows in size, so does the amount of text based information that exists. For many application spaces it is paramount to isolate and identify texts that relate to a particular topic. While one-class classification would be ideal for such analysis, there is a relative lack of research regarding efficient approaches with high predictive power. By noting that the range of documents we wish to identify can be represented as positive linear combinations of the Vector Space Model representing our text, we propose Conical classification, an approach that allows us to identify if a document is of a particular topic in a computationally efficient manner. We also propose Normal Exclusion, a modified version of Bi-Normal Separation that makes it more suitable within the one-class classification context. We show in our analysis that our approach not only has higher predictive power on our datasets, but is also faster to compute.

\end{abstract}

\section{Introduction}

In the era of the rapid development of computers and the Internet, information on a wide range of topics is pervasive. The amount of text based data is ever increasing in size, magnitude, and variety. Whether it is for e-commerce \cite{Xiao_2021}, clinical diagnosis determination \cite{le2021machine}, or fake news detection \cite{ahmed2018detecting} it is vital to have efficient mechanisms for topic classification in order to effectively parse and process text based media.


Most of the research on topic classification uses these implementations within a binary classification or multi-class classification context \cite{trstenjak2014knn, zhang2011comparative, kim2019research, kim2019multi, liu2018research}. Comparatively, there is a relative dearth of content variety discussing and proposing different algorithms that can identify text on a particular subject from a variety of subjects in a One-Vs-All configuration, especially regarding how to use vector representations of documents with low computational costs. This is unfortunate, as one class classification of text enables us to identify text of a particular form from a potentially non-exhaustible set of potential topics. In such a setting, it would be arduous to identify all potential topics we may come across and extremely time-consuming to label enough data to train a model for multi-class classification.

In practice, the lack of research into one class topic determination has lead to subpar implementations for the sake of speed. One of the best examples of the ramifications of this lack of research focus is insider threat detection systems. Despite insider threat detection primarily working with log and textual information, the vast majority of published work on the subject do not utilize Natural Language Processing in their implementations \cite{wei2021insider, tuor2017deep, meng2018deep, le2018benchmarking, le2019machine}. Many that do simply sum over TF-IDF vectors before feeding the result as a feature into detection models \cite{chattopadhyay2018scenario, sajjanhar2019image}.

We aim to tackle these issues head on. Our contributions are as follows:

\begin{itemize}
  \item We propose Normal Exclusion, a re-framing of Bi-Normal Separation enabling usage for one-class classification.
  \item We show that our approach, Conical Classification (CC), achieves optimal performance when compared to alternative one-class topic determination strategies.
\end{itemize}

\section{Related Work}
With the intention of assessing the predictive power of one-class based text classification methods, Joffe et al. has compared one-class support vector machines (OCSVM) to binary support vector machines (SVM) to identify specific phenotypes in breast cancer. They found that OCSVM performed comparably to SVM in balanced dataset problem spaces and outperformed SVM in highly imbalanced datasets \cite{joffe2015expert}.

Zhuang et al. concurs, citing the improved performance of switching from a SVM to OCSVM approach for minority class classification. They use a general framework which first uses the minority class for training in the one-class classification stage, then incorporate data from the majority class to improve the generalization performance of the constructed classifier \cite{zhuang2006parameter}.

It turns out that OCSVMs have a wide adoption rate for one-class text classification problems. Additional examples include Manevitz et al. using them for document classification \cite{manevitz2001one}, and Seo utilizing a OCSVM to help classify images in a database using color and text content for content-based image retrieval \cite{seo2007application}. 

Ensemble based methodologies have been used in practice as well. Hempstalk et al. has utilized an ensemble-based approach using C4.5 decision trees with Laplace smoothing to isolate real target values from those of an artificial class \cite{hempstalk2008one}, validating performance on various UCI datasets as well as a custom typist dataset. Anderka et al. utilized a similar approach to detect text quality flaws, using a Random Forest as the base classifier instead \cite{anderka2011detection}. Unfortunately, despite their higher memory and computation requirements, such approaches have little performance benefits compared to the OCSVM; Hempstalk et al.'s results indicated that their ensemble approach was not demonstratively superior to the OCSVM approach.

While not traditional one-class classification algorithms, there are a set of classifiers that co-train using a set of positive labeled data as well as a set of unlabeled data for evaluation. Denis et al. has developed the Positive Naive Bayes (PNB) classifier that works under this setting, using it successfully to classify documents in the 20-Newsgroup dataset \cite{denis2003text}.

One-class topic determination is a problem space where it is paramount to be computationally fast with low resources in order to process large numbers of documents in a short amount of time. This has traditionally excluded recent advancements in Natural Language Processing such as embeddings from the discussion, as these take significant amounts of computation time on the modest hardware such application spaces necessitate. This has resulted in very few publications dedicated to assessing their application to the space. Ruff et al. propose Context Vector Data Description (CVDD) \cite{ruff2019self}, a textual anomaly detection algorithm that builds upon word embedding models to learn multiple sentence representations that capture multiple semantic contexts via the self-attention mechanism. Hu et al. extended uni-modal Support Vector Data Description (SVDD) to a multiple modal one, building Multi-modal Deep Support Vector Data Description (mSVDD) with multiple hyperspheres, enabling them to build better descriptions for target one-class data \cite{hu-etal-2021-one}.

The methodology used to create the vector representations of documents can be just as important as the detection algorithm used. One main approach that has come about as a result is term frequency (TF) – inverse document frequency (IDF). TF-IDF is the product of two statistics: TF and IDF. TF, as its name suggests, refers to the normalized frequency $f$ of a word $w_j$ that appears in the given document $D$. Originally coined as term specificity by Jones \cite{jones1972statistical}, IDF provides a measure of how much information a word provides depending on how common the word is in a given corpus.

TF-IDF has been successfully used for topic classification in a variety of scenarios, ranging from social media \cite{lee2011twitter}, research analysis \cite{kim2019research}, and news discovery \cite{hakim2014automated}. As a result, much research has been done on modifications to improve performance. Martineau et. al. has proposed Delta TF-IDF which scales weights using word scores before classification and boasts a higher accuracy than standard TF-IDF \cite{martineau2009delta}. Forman studies replacing TF-IDF with Bi-Normal Separation (BNS), eliminating the need for fine-tuned feature selection and performs exceptionally well on short length documents \cite{forman2003extensive}. Domeniconi et. al. used a supervised variant to prevent the IDF term from affecting documents within the category under analysis, so that terms frequently appearing in said category are not penalized \cite{domeniconi2015comparison}. 

More recently, vector representations have been developed that use embeddings, such as BERT \cite{devlin2018bert} and GloVe \cite{pennington2014glove}. Such embeddings allow for words with similar meanings to have a similar representation which has allowed for the impressive performance of deep learning methods on complex and intricate natural language processing problem spaces.

\section{Normal Exclusion}
BNS, which is the measure of how much the probability of occurrence of a given word in the positive class differs from the probability of occurrence of a given word in the negative class, has a couple of key benefits as a VSM metric: it is excellent at ranking words for automated feature selection filtering, it has the best performance in single metric VSM analyses, and is  consistently a member of the optimal pairs of VSM metrics Forman et al. evaluated \cite{forman2008bns}. Thus, being able to utilize BNS within a one-class context would be ideal.

The formula used to calculate BNS is given in Equation 1. Here, tpr is the true positive rate $P(word | positive class)$ as determined via $\frac{tp}{pos}$, where $tp$ is the number of positive training cases containing the word and $pos$ is the number of positive training cases. Likewise, $fpr$ refers to the false positive rate $P( word | negative class)$ as determined via $\frac{fp}{neg}$, where $fp$ is the number of negative training cases containing word and neg is the number of negative training cases. $F^{-1}$ is the inverse  Normal cumulative distribution function. $\epsilon$ is a number with small magnitude added to avoid the undefined scenario of $F^{-1}(0)$; for the purposes of our analysis, we set $\epsilon$ to $0.0005$, or half a count out of 1000.

\vspace{-4mm}
\begin{equation}
    \resizebox{0.75\linewidth}{!}
    {$
    BNS = \left|F^{-1}(tpr + \epsilon) - F^{-1}(fpr + \epsilon)\right|
    $}
\end{equation}

A naive translation to a one-class regimen would be to merely remove BNS's dependence on the $fpr$ term. Thus, each word would be scaled in relation to its frequency of occurrence within our positive training set. This leads to issues, as words with a naturally high occurrence in language such as the, be, to, of, a, etc. will have predominantly high scaled values. One may try to work around these effects by removing stopwords and unrelated words from our corpus, but this can require significant hand-tuning by an expert in the field while increasing overhead computation costs.

We propose an alternative solution that takes advantage of the nature of one-class classification, recalling that we wish to be able to identify text of a particular topic from any assortment of topics possible from the language. We simply need to estimate the $fpr$ of the word with the frequency of the word in our given language. For English, there are large corpuses from which we can extract this information, for example the Oxford English Corpus (OEC) is a dataset that presents all types of English, from blogs to newspaper articles to literary novels and even social media, sourcing from Englishes from the United Kingdom, the United States, Ireland, Australia, New Zealand, the Caribbean, Canada, India, Singapore, and South Africa. For our purposes, we compiled the frequencies of the top $\frac{1}{3}$ million words in the human language using Tatman's English word count dataset \cite{tatmankaggleword, brantsgooglewordset} and stored them within a dictionary for rapid lookup.

We can safely set the frequencies of words that do not appear in our dictionary to $0$, as these include words that rarely appear in standard language; such words include abaptiston, abaxile, grithbreach, gurhofite, zarnich, and zeagonite. Indeed, according to Oxford's compiled statistics, the combined frequency of occurence for all such words is approximately a percent of the entire lexicon of the English language, easily within the margin of error for our analysis \cite{oecstats}.

\vspace{-0.5cm}
\begin{equation}
    \resizebox{0.8\linewidth}{!}
    {$
    NE = \left|F^{-1}(tpr + \epsilon) - F^{-1}(Dict[word] + \epsilon)\right|
    $}
\end{equation}

We coin our tweaked formula Normal Exclusion (NE), as it excludes, or reduces, the weightage of words that are inconsequential to determining the topic of text without requiring a negative corpus to be present. The formula for NE is shown in Equation 2. Here, Dict[word] represents the frequency value for the given word as found within our dictionary.

We will scale NE by TF for our model developing the NE-TF VSM. Our representation of a word in a model will thus be determined by how frequently a word occurs in our corpus, scaled by the statistical significance of the word within the evaluated text. Higher magnitude values give a strong indication that the vector is about our target topic, while lower values would lead to a lower confidence that such a conclusion is correct.

\section{Conical Classification}
\subsection{Why Positive Span}
VSM is based on the notion of vector similarity; the model assumes that the relevance of a document to another document is roughly equal to the document-query similarity. Under this model, the documents are represented using the \textit{bag-of-words} approach. This means that documents are translated to $n$-dimensional vectors, where each dimension corresponds to a word based on a compiled set of terms known as a vocabulary. Under such models, we map a given topic to a certain subset of the compiled vocabulary.

It is not enough however for a document to have a high frequency of words included within the subset to be classified as a given topic. Combinations of words are vital to the classification process. For a timely example, a news article regarding COVID-19 and an administration protocol manual on COVID-19 vaccines will both strongly correlate to words such as vaccines, dosages, Pfizer, Moderna, among others. To distinguish between these two topics, we would need contextual words such as policy, mandate, and president to identify a news article, and words like intramuscular, angle, deltoid, and subcutaneous would likely exist within an administration protocol manual. While these contextual words will have a lower correlation to a given topic, they are nonetheless paramount for an effective classification model. This leads to a high significance of vector orientation within a VSM as it is crucial to keep track of how a word represented by a certain dimension relates to words represented by different dimensions.

The high interdependence between VSMs and orientation allows one to assess document similarity solely from the context of vector angles. For example, to rank similarity within a category, a simple and popular mechanism is to calculate the Relevance Status Value which computes the cosine of the angle between the query and each document in the collection \cite{rao2018computational}. The larger the cosine value, the smaller the angle, and the more similar the documents being compared are. It is important to note at this point that while vector magnitude would typically be a crucial metric to consider as well, Rao et al. furthers, stating VSM vectors are typically normalized before further computation and analysis is done.

\begin{figure}[h]
\centerline{\includegraphics[width=\linewidth,height=\textheight,keepaspectratio]{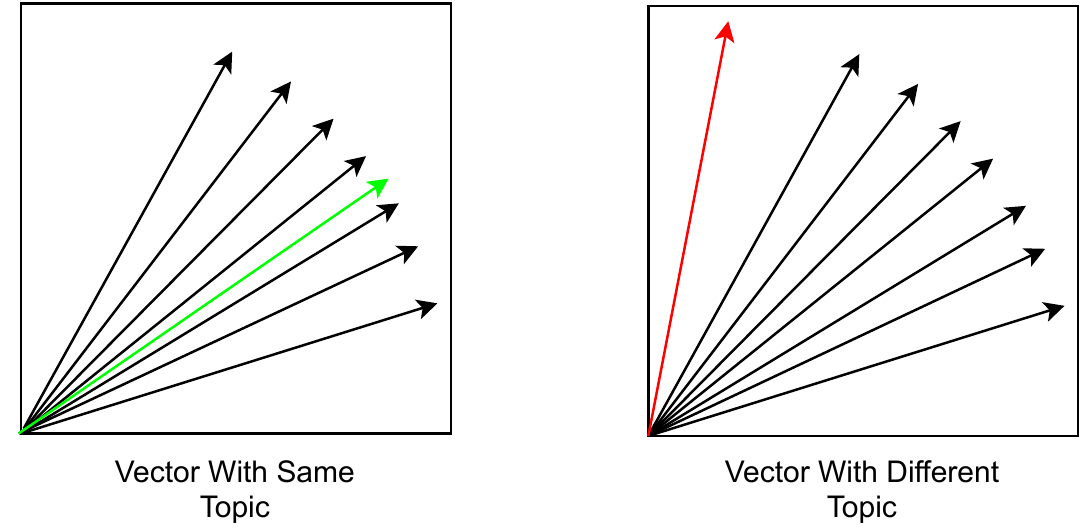}}
\caption{New document vector of the same topic versus new document vector of a different topic. Green refers to a document that will be classified as having the same topic, red will be classified as not having the same topic.}
\label{fig}
\end{figure}

This means that documents of the same topic will have smaller angles between each other than those comprised of different topics altogether. Extrapolating from this observation to the comparison of a document to an entire corpus, we expect for vectors corresponding to the same topic to be close to the center of the distribution of corpus vectors in order to have a low angle to all vectors in the corpus. Similarly, we expect vectors from a different topic to have a high angle from the vectors in the corpus. Figure 1 provides an illustration of the expected phenomenon.

\begin{figure}[h]
\centerline{\includegraphics[width=\linewidth,height=\textheight,keepaspectratio]{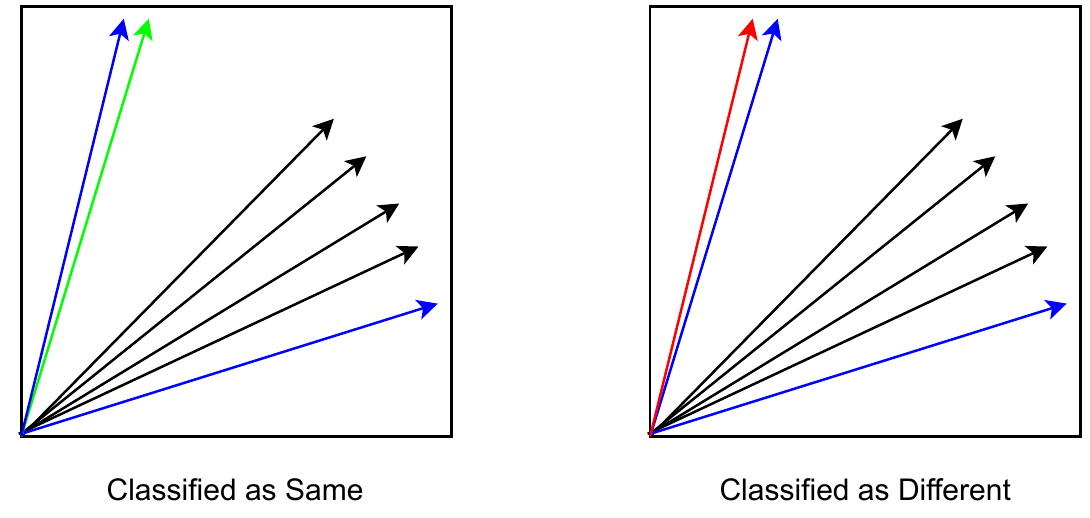}}
\caption{Edge cases for our classification system. Green and red remain as defined in Figure 1. Blue refers to our corpus fringe vectors.}
\label{fig}
\end{figure}

Note we do not yet consider documents that are edge case scenarios. To simplify nomenclature for further discussion, we refer to vectors within our corpus that are most dissimilar to the other vectors in the corpus our fringe vectors. We consider fringe vectors to be as distant from the corpus as possible while still being considered as having the same topic. Thus, as shown in Figure 2, the similarity with respect to a fringe vector is not sufficient to be classified as having the same topic as the given corpus; if a vector is similar to a fringe vector, but less similar to rest of the corpus than the fringe vector, we will consider the vector being evaluated to be of a different topic. In other words, a vector must be in-between our fringe vectors across all dimensions to be considered as having the same topic as our corpus. 

\begin{figure}[h]
\centerline{\includegraphics[width=0.3\linewidth,height=\textheight,keepaspectratio]{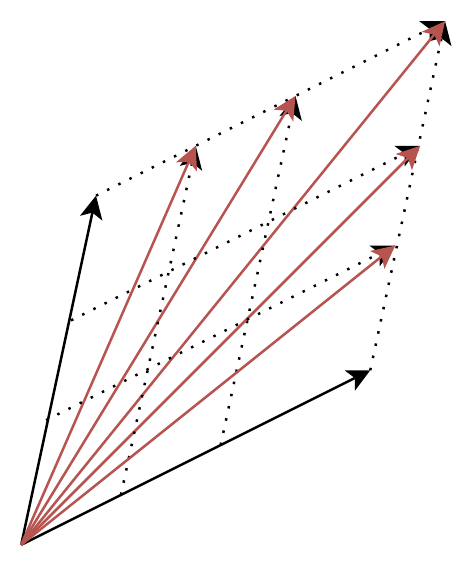}}
\caption{Any vector found between two vectors can be created from the linear combination of its surrounding vectors.}
\label{fig}
\end{figure}

From here, we can translate the classification problem into a linear combination problem. As shown in Figure 3 for the two dimensional case, any vector found in between two vectors can be represented by their linear combination. We define a vector as being in-between two vectors if the sum of its angles to each vector is equal to the angle between the two vectors themselves and it lies on the plane defined by the two vectors. Note this vector can always be calculated as a linear combination of its surrounding vectors; Algorithm 1 shows an approach based on binary search that allows one to identify the scalar combinations needed to recreate the target vector. Here, $cos_{sim}$ refers to cosine similarity \cite{sitikhu2019comparison}, $target$ is the vector we are trying to recreate, $x$ and $y$ are the vectors $target$ is in-between while $\lambda_x$ and $\lambda_y$ are the scalar values such that $x\lambda_x + y\lambda_y = target$.

\begin{algorithm}[htbp]
\SetAlgoLined
\KwResult{$\lambda_x$, $\lambda_y$}
 $vector_{one} = x$\;
 $vector_{two} = y$\;
 $mid = \frac{vector_{one} + vector_{one}}{2}$\;
 $\lambda_x = \frac{1}{2}$\;
 $\lambda_y = \frac{1}{2}$\;
 $level = 1$\;
 \While{$mid \neq target$}{
  $level = level + 1$\;
  $sim_{one} = cos_{sim}(vector_{one}, target)$\;
  $sim_{two} = cos_{sim}(vector_{two}, target)$\;
  \eIf{$sim_{one} \geq sim_{two}$}{
   $mid = vector_{two}$\;
   $\lambda_x = \lambda_x + 2^{-level}$\;
   $\lambda_y = \lambda_y - 2^{-level}$\;
   }{
   $mid = vector_{one}$\;
   $\lambda_x = \lambda_x - 2^{-level}$\;
   $\lambda_y = \lambda_y + 2^{-level}$\;
  }
 }
 \caption{Binary Search Approach To Finding Linear Combination Scalars For Target Vector In-between Two Vectors}
\end{algorithm}

This conclusion also makes intuitive sense. As discussed earlier, we can identify a document as being from a particular topic if it has word combinations that indicate as such. A vector that is a linear combination of those within the corpus must have one or more such identifying word combinations as a result.

It is important to note that by linear combinations, we specifically refer to the set of positive linear combinations. As mentioned earlier, orientation of vectors is crucial in regards to which documents and word combinations they represent. A negatively scaled vector represents the complete opposite document than a positively scaled counterpart and thus should not be used for topic classification.

We have shown it is enough to compose a vector as a positive linear combination of the vectors in a corpus to confirm that it is regarding a similar topic. In other words, a document has the same topic as a corpus if its vector representation is within the positive span of the corpus.

\subsection{Conical Sets}

\begin{figure*}[htbb]
    \centering
    \def\svgwidth{\textwidth}
    \centerline{\includegraphics[width=\linewidth,height=\textheight,keepaspectratio]{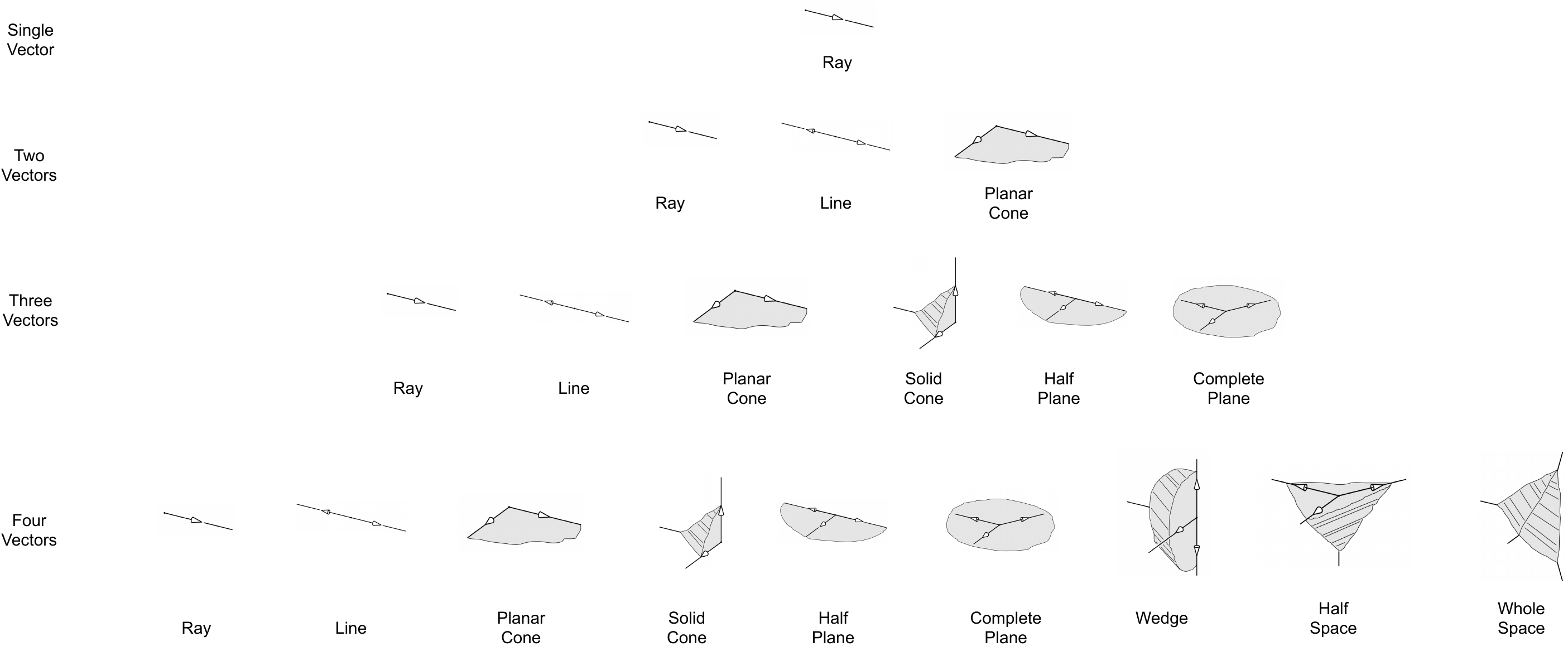}}
    \caption{Conical set subspaces comprised of different vector totals in three dimensional space.}
    \label{fig:some_label}
\end{figure*}

The positive span of vectors $v_1$ through $v_k \in \mathbb{R}^n$ is the linear combination $\sum_i^k\lambda_iv_i$ where $\lambda_i \geq 0$ for all i = 1, ..., k \cite{davis1954theory}. Note that the original definition made by Davis allows for the zero vector to be included within the positive span. However, the zero vector within the VSM context represents a vector with none of the terms corresponding to the corpus topic; we thus wish to exclude the zero vector from our span in order to properly classify documents based on their topic. Our new span, coined the conical span, of vectors $v_1$ through $v_k \in \mathbb{R}^n$ is the linear combination $\sum_i^k\lambda_iv_i$ where $\sum_i^k \lambda_i > 0$ for all $i = 1, ..., k$.

We define the conical set that can be defined via the conical span of a finite number of vectors in Equation 3.

\vspace{-4mm}
\begin{equation}
    \resizebox{0.85\linewidth}{!}
    {$
        conical(S) := \left\{\lambda_1v_1 + ... + \lambda_kv_k : \lambda_1 + ... + \lambda_k > 0 \right\}
    $}
\end{equation}

Conical span enables a large range of possibilities from the positive span of vectors; Figure 4 showcases the vast representational power in three dimensional space, where the addition of extra vectors dramatically increases the variety of subspace shapes that can be created \cite{positivespanchart}.

\subsection{Two Vector Comparison}
At this point, we have shown that it is sufficient for topic classification that a vector is within the conical span, and we have displayed the expressive power of the conical span. We will now go over an efficient mechanism to determine if a vector is within the conical span. As Rao et al. claims is standard for VSM vectors, we assume all vectors in our corpus as well as our evaluation vectors are unit norm in length \cite{rao2018computational}.

In order to train our CC system, we simply find the largest value and the smallest value for every dimension in our corpus vectors, and store them within two vectors for analysis later on. Then when it comes to predicting with CC, we simply need to compare our evaluation vector with both vectors in order to determine if the vector belongs to our corpus.

We prove this claim via the following lemmas and theorems.

\begin{lemma}
There is no unit vector within the conical span that is larger in one or more dimensions than the max vector or smaller in one or more dimensions than the min vector.
\end{lemma}
\begin{proof}
We prove by contradiction. Assume there is a vector in the conical span whose value in one or more dimensions is larger than than the max vector or smaller than the min vector. The values in the max and min vectors are set by the fringe vectors for the given dimensions, due to these vectors having the largest deviation from the corpus acceptable. For a vector to have a value outside of this range, the given vector must deviate further from the rest of the corpus than our fringe vectors. This leads to a contradiction; by definition, any vector less similar to our corpus than the fringe vectors must not be classified as being of the same topic and thus within the conical span.
\end{proof}

\begin{lemma}
There is no unit vector outside the conical span that is smaller in a given dimension than the max vector and larger in a given dimension than than the min vector.
\end{lemma}
\begin{proof}
We prove by contradiction. Assume that a unit length vector outside the conical span exists such that its values are in between the min and max vectors. As mentioned in Lemma 4.1, the max and min vectors are defined by the value of our fringe vectors for each dimension of our VSM. A value closer in similarity to the rest of our corpus is by definition within our conical span. This leads to a contradiction: a vector cannot both be more similar to main vectors within our corpus than our fringe vectors and be classified as a different topic. 
\end{proof}

\begin{theorem}
All possible unit vectors in the conical span can be represented by a max and min vector.
\end{theorem}
\begin{proof}
By combining Lemmas 4.1 and 4.2, we arrive at the conclusion that Theorem 4.3 is indeed correct.
\end{proof}

This result enables us to rapidly train and determine if a given vector is of a certain topic or not. If a vector is not the zero vector, is less than the max vector across all dimensions, and is greater than the min vector across all dimensions, then we classify the vector as being of the same topic as it is within the conical span of the topic training corpus.

\section{Evaluation Methodology}

\subsection{Baseline Models}
As detailed in the Related Works section, OCSVMs have an extremely high adoption rate within the space, thus for our analysis, we evaluate the performance of OCSVMs on the following kernel functions: linear, sigmoid, radial basis function (RBF), and polynomial (Poly).

To represent our set of ensembles, we will train a One Class Random Forest classifier (OCRF) using Goix et al.'s splitting method \cite{goix2017one} as well as an Isolation Forest classifier (IsolFor) \cite{liu2008isolation}. For both methods, we will use 1000 estimators.

We also utilize PNB as a baseline measure. Since we wish to evaluate its performance in the one-class classification regime, we will use the evaluation data itself as the unlabeled set of data for training the algorithm; this allows us to only pass in the positive set of data points during training as is the case for traditional one-class classification algorithms.

Finally, to represent embedding based models, we use CVDD as our representation for context preserving embedding based approaches as well as for neural NLP, taking advantage of the official implementation known as CVDD-PyTorch \cite{ruff2019self}. Both GloVe and BERT models are assessed for evaluation purposes, with embedding size, attention size, and number of attention heads set to be the best performing configuration .

Except for our CVDD baselines, all of our baseline models will use TF-IDF as the VSM of choice.

\subsection{Dataset}
Our intent is to evaluate our baselines as well as CC in scenarios that can require high performance. As mentioned in the Introduction, one main place where this can occur is in insider threat detection. The golden standard dataset for insider threats is the CERT Insider Threat dataset, the largest public repository of insider threat scenarios compiled after analyzing 1,154 actual insider incidents \cite{glasser2013bridging}. Within this dataset, there are three key website topics that are crucial to detect: Keylogger websites, Wikileaks-like websites, and job posting sites. We extract the text related to both Keylogging and Wikileaks by hand-labeling the text content within version 4.2 in order to use them both for evaluation purposes.

For the purpose of evalauting the latter of the three, we extract text related information from Vidros et al.'s Fake JobPosting Prediction dataset \cite{vidros2017automatic}, and from PromptCloud's job dataset \cite{monsterjobs17}. Both are high quality datasets listing full descriptions of jobs with large varieties, and versions of both datasets have been used by a plethora of publications \cite{balachander2018ontology, kim2019fraud, alghamdi2019intelligent, mahbub2018using, reddy2018analysis}. For our purposes, we extract text data from the real job postings in Vidros et al.'s dataset.

We also desire our evaluation set to have exposure to e-commerce applications, medical record information, and fake news articles. For our ECommerce dataset, we utilize the Women’s Clothing E-Commerce dataset \cite{agarap2018statistical}, which has seen popularity for sentiment analysis and text classification tasks \cite{sun2019sentiment, lin2020sentiment, koustalocal, cascaro2019aggregating}. Our MedicalTranscription dataset consists of text extracted from the Collection of Transcribed Medical Transcription Sample Reports and Examples \cite{medtransdataset}, a dataset of interest in academia both from a natural language processing perspective as well as from a medical assessment point of view \cite{beattie1994separation, moramarco2021towards, zuccon2014identification}. Finally, for FakeNews, we utilize the Fake and real news dataset \cite{ahmed2018detecting, ahmed2017detection}; this dataset is especially relevant due to recent increases in the proliferation and rapid diffusion of fake news on the Internet.

We chose this set of classification markers not only due to its representation of some of the fields we expect CC to be applicable, but also due to the high variability in text length and composition; our Wikileaks and Keylogger datasets are small length texts composed primarily of keywords, whereas the MoviePlot and MedicalTranscription datasets have relatively verbose text covering complex and protean topic ranges. This large variety is crucial as research has shown that text length and topic variations have a dramatic affect on text-based classification performance \cite{wang2012baselines}.

\begin{table*}[htbp]
\vspace{-4mm}
\caption{Performance metrics and computation times for all algorithms on evaluation datasets. Best results are bolded.}
\centering
\resizebox{\linewidth}{!}{%
\begin{tabular}{|c|c|c|c|c|c|c|c|}
\hline
Dataset                    & Model        & Accuracy & Balanced Accuracy             & Precision            & Recall & F1 Score & Time (s) \\ \specialrule{.1em}{.05em}{.05em}

\multirow{8}{*}{ ECommerce } & Linear OCSVM & $0.900 \pm 0.001$ & $0.755 \pm 0.003$ & $0.978 \pm 0.002$ & $0.512 \pm 0.007$ & $0.672 \pm 0.006$ & $183.255 \pm 3.821$ \\ \cline{2-8} 
                             & RBF OCSVM & $0.899 \pm 0.001$ & $0.753 \pm 0.004$ & $0.981 \pm 0.003$ & $0.508 \pm 0.008$ & $0.669 \pm 0.007$ & $201.339 \pm 3.792$\\ \cline{2-8} 
                             & Sigmoid OCSVM & $0.899 \pm 0.002$ & $0.752 \pm 0.005$ & $0.979 \pm 0.003$ & $0.508 \pm 0.011$ & $0.668 \pm 0.009$ & $193.402 \pm 7.382$\\ \cline{2-8} 
                             & Poly OCSVM & $0.885 \pm 0.006$ & $0.712 \pm 0.002$ & $\bm{ 0.995 \pm 0.001 }$ & $0.424 \pm 0.003$ & $0.595 \pm 0.003$ & $183.813 \pm 2.639$\\ \cline{2-8} 
                             & IsolFor & $0.199 \pm 0.000$ & $0.500 \pm 0.000$ & $0.199 \pm 0.000$ & $\bm{ 1.000 \pm 0.000 }$ & $0.333 \pm 0.000$ & $280.893 \pm 12.543$\\ \cline{2-8} 
                             & OCRF & $0.953 \pm 0.000$ & $0.947 \pm 0.0176$ & $0.800 \pm 0.000$ & $0.898 \pm 0.036$ & $0.889 \pm 0.000$ & $189.252 \pm 244.041$\\ \cline{2-8} 
                             & PNB & $0.638 \pm 0.202$ & $0.762 \pm 0.092$ & $0.786 \pm 0.203$ & $0.757 \pm 0.305$ & $0.687 \pm 0.153$ & $111.839 \pm 59.581$\\ \cline{2-8}
                             & GloVe CVDD & $0.948 \pm 0.017$ & $ 0.906 \pm 0.026$ & $ 0.951 \pm 0.015$ & $ 0.983 \pm 0.007$ & $ 0.967 \pm 0.011$ & $ 188.462 \pm 12.773$\\ \cline{2-8}
                             & BERT CVDD & $0.951 \pm 0.128$ & $ 0.910 \pm 0.013$ & $ 0.954 \pm 0.017$ & $ 0.987 \pm 0.012$ & $ 0.973 \pm 0.028$ & $ 233.626 \pm 23.796$\\ \cline{2-8}
                             & CC & $\bm{ 0.988 \pm 0.009 }$ & $\bm{ 0.988 \pm 0.007 }$ & $0.956 \pm 0.005$ & $0.988 \pm 0.009$ & $\bm{ 0.971 \pm 0.002 }$ & $\bm{ 10.878 \pm 0.036 }$\\ \specialrule{.1em}{.05em}{.05em}

\multirow{8}{*}{ FakeNews } & Linear OCSVM & $0.818 \pm 0.039$ & $0.685 \pm 0.082$ & $0.819 \pm 0.157$ & $0.430 \pm 0.232$ & $0.495 \pm 0.175$ & $496.206 \pm 5.353$\\ \cline{2-8} 
                            & RBF OCSVM & $0.813 \pm 0.033$ & $0.706 \pm 0.065$ & $0.772 \pm 0.184$ & $0.500 \pm 0.229$ & $0.538 \pm 0.086$ & $533.529 \pm 11.002$\\ \cline{2-8} 
                            & Sigmoid OCSVM & $0.760 \pm 0.046$ & $0.648 \pm 0.091$ & $0.757 \pm 0.259$ & $0.436 \pm 0.347$ & $0.389 \pm 0.166$ & $514.913 \pm 19.340$\\ \cline{2-8} 
                            & Poly OCSVM & $0.850 \pm 0.014$ & $0.722 \pm 0.053$ & $0.846 \pm 0.071$ & $0.476 \pm 0.128$ & $0.592 \pm 0.091$ & $470.485 \pm 2.950$\\ \cline{2-8} 
                            & IsolFor & $0.238 \pm 0.000$ & $0.500 \pm 0.000$ & $0.238 \pm 0.000$ & $\bm{ 1.000 \pm 0.000 }$ & $0.384 \pm 0.000$ & $278.308 \pm 5.449$\\ \cline{2-8} 
                            & OCRF & $0.930 \pm 0.000$ & $0.955 \pm 0.000$ & $0.761 \pm 0.000$ & $1.000 \pm 0.000$ & $0.864 \pm 0.000$ & $197.232 \pm 255.352$\\ \cline{2-8} 
                            & PNB & $0.624 \pm 0.223$ & $0.773 \pm 0.097$ & $0.900 \pm 0.142$ & $0.697 \pm 0.283$ & $0.729 \pm 0.152$ & $218.296 \pm 38.515$\\ \cline{2-8}
                             & GloVe CVDD & $0.906 \pm 0.003$ & $ 0.881 \pm 0.011$ & $ 0.938 \pm 0.018$ & $ 0.936 \pm 0.022$ & $ 0.936 \pm 0.002$ & $ 282.293 \pm 31.88$\\ \cline{2-8}
                             & BERT CVDD & $0.899 \pm 0.121$ & $ 0.878 \pm 0.218$ & $ 0.923 \pm 0.342$ & $ 0.933 \pm 0.231$ & $ 0.927 \pm 0.153$ & $ 322.513 \pm 49.659$\\ \cline{2-8}
                            & CC & $\bm{ 0.985 \pm 0.005 }$ & $\bm{ 0.985 \pm 0.004 }$ & $\bm{ 0.955 \pm 0.002 }$ & $0.987 \pm 0.006$ & $\bm{ 0.969 \pm 0.001 }$ & $\bm{ 13.952 \pm 0.026 }$\\ \specialrule{.1em}{.05em}{.05em}  

\multirow{8}{*}{ Jobs } & Linear OCSVM & $0.913 \pm 0.001$ & $0.768 \pm 0.004$ & $0.971 \pm 0.002$ & $0.540 \pm 0.009$ & $0.694 \pm 0.007$ & $368.263 \pm 6.227$\\ \cline{2-8} 
                        & RBF OCSVM & $0.910 \pm 0.001$ & $0.758 \pm 0.003$ & $0.975 \pm 0.002$ & $0.520 \pm 0.007$ & $0.678 \pm 0.006$ & $408.096 \pm 5.810$\\ \cline{2-8} 
                        & Sigmoid OCSVM & $0.912 \pm 0.006$ & $0.766 \pm 0.001$ & $0.968 \pm 0.001$ & $0.537 \pm 0.003$ & $0.690 \pm 0.002$ & $386.841 \pm 15.345$\\ \cline{2-8} 
                        & Poly OCSVM & $0.903 \pm 0.001$ & $0.736 \pm 0.002$ & $\bm{ 0.989 \pm 0.001 }$ & $0.474 \pm 0.005$ & $0.641 \pm 0.005$ & $364.713 \pm 6.367$\\ \cline{2-8} 
                        & IsolFor & $0.181 \pm 0.000$ & $0.500 \pm 0.000$ & $0.181 \pm 0.000$ & $\bm{ 1.000 \pm 0.000 }$ & $0.307 \pm 0.000$ & $291.792 \pm 8.304$\\ \cline{2-8} 
                        & OCRF & $0.961 \pm 0.000$ & $0.976 \pm 0.000$ & $0.818 \pm 0.000$ & $1.000 \pm 0.000$ & $0.900 \pm 0.000$ & $196.165 \pm 253.234$\\ \cline{2-8} 
                        & PNB & $0.585 \pm 0.214$ & $0.696 \pm 0.154$ & $0.684 \pm 0.290$ & $0.869 \pm 0.214$ & $0.690 \pm 0.166$ & $189.128 \pm 58.950$\\ \cline{2-8}
                             & GloVe CVDD & $0.896 \pm 0.037$ & $ 0.836 \pm 0.056$ & $ 0.910 \pm 0.054$ & $ 0.961 \pm 0.033$ & $ 0.933 \pm 0.025$ & $ 271.637 \pm 30.758$\\ \cline{2-8}
                             & BERT CVDD & $0.886 \pm 0.023$ & $ 0.831 \pm 0.049$ & $ 0.903 \pm 0.034$ & $ 0.958 \pm 0.025$ & $ 0.918 \pm 0.012$ & $ 316.066 \pm 32.659$\\ \cline{2-8}
                        & CC & $\bm{ 0.995 \pm 0.017 }$ & $\bm{ 0.994\ \pm 0.000\ }$ & $0.985\ \pm 0.006$ & $0.993\ \pm 0.004$ & $\bm{ 0.988 \pm 0.004 }$ & $\bm{ 11.115 \pm 0.021 }$\\ \specialrule{.1em}{.05em}{.05em}  

\multirow{8}{*}{ Keylogger } & Linear OCSVM & $0.999 \pm 0.004$ & $0.706 \pm 0.041$ & $1.000 \pm 0.000$ & $0.413 \pm 0.081$ & $0.580 \pm 0.078$ & $\bm{ 10.330 \pm 1.035 }$\\ \cline{2-8} 
                             & RBF OCSVM & $0.999 \pm 0.009$ & $0.705 \pm 0.080$ & $1.000 \pm 0.000$ & $0.410 \pm 0.161$ & $0.564 \pm 0.156$ & $11.373 \pm 0.366$\\ \cline{2-8} 
                             & Sigmoid OCSVM & $0.999 \pm 0.001$ & $0.705 \pm 0.093$ & $1.000 \pm 0.000$ & $0.411 \pm 0.187$ & $0.556 \pm 0.192$ & $10.606 \pm 0.766$\\ \cline{2-8} 
                             & Poly OCSVM & $0.999 \pm 0.007$ & $0.745 \pm 0.066$ & $1.000 \pm 0.000$ & $0.491 \pm 0.131$ & $0.648 \pm 0.124$ & $10.361\ \pm 0.831$\\ \cline{2-8} 
                             & IsolFor & $0.791 \pm 0.294$ & $0.720 \pm 0.031$ & $0.666 \pm 0.470$ & $0.649 \pm 0.251$ & $0.428 \pm 0.303$ & $276.514 \pm 22.813$\\ \cline{2-8} 
                             & OCRF & $1.000 \pm 0.000$ & $1.000 \pm 0.000$ & $0.999 \pm 0.000$ & $1.000 \pm 0.000$ & $0.999 \pm 0.000$ & $162.149 \pm 199.354$\\ \cline{2-8} 
                             & PNB & $0.274 \pm 0.437$ & $0.636 \pm 0.219$ & $0.268 \pm 0.440$ & $1.000 \pm 0.000$ & $0.271 \pm 0.439$ & $74.266 \pm 74.840$\\ \cline{2-8}
                             & GloVe CVDD & $1.000 \pm 0.000$ & $ 1.000 \pm 0.000$ & $ 1.000 \pm 0.000$ & $ 1.000 \pm 0.000$ & $ 1.000 \pm 0.000$ & $ 259.308 \pm 13.955$\\ \cline{2-8}
                             & BERT CVDD & $1.000 \pm 0.000$ & $ 1.000 \pm 0.000$ & $ 1.000 \pm 0.000$ & $ 1.000 \pm 0.000$ & $ 1.000 \pm 0.000$ & $ 319.108 \pm 17.433$\\ \cline{2-8}
                             & CC & $\bm{ 1.000 \pm 0.000 }$ & $\bm{ 1.000 \pm 0.000 }$ & $\bm{ 1.000 \pm 0.000 }$ & $\bm{ 1.000 \pm 0.000 }$ & $\bm{ 1.000 \pm 0.000 }$ & $15.610 \pm 0.055$\\ \specialrule{.1em}{.05em}{.05em}  

\multirow{8}{*}{ MedicalTranscriptions } & Linear OCSVM & $0.971 \pm 0.006$ & $0.743 \pm 0.007$ & $0.709 \pm 0.012$ & $0.496 \pm 0.015$ & $0.583 \pm 0.012$ & $77.501 \pm 1.400$\\ \cline{2-8} 
                                         & RBF OCSVM & $0.973 \pm 0.007$ & $0.734 \pm 0.011$ & $0.766 \pm 0.010$ & $0.475 \pm 0.023$ & $0.586 \pm 0.019$ & $94.011 \pm 2.294$\\ \cline{2-8} 
                                         & Sigmoid OCSVM & $0.971 \pm 0.006$ & $0.741 \pm 0.008$ & $0.707 \pm 0.012$ & $0.492 \pm 0.016$ & $0.578 \pm 0.012$ & $83.649 \pm 4.104$\\ \cline{2-8} 
                                         & Poly OCSVM & $0.973 \pm 0.002$ & $0.667 \pm 0.002$ & $ 0.981 \pm 0.005 $ & $0.335 \pm 0.005$ & $0.499 \pm 0.005$ & $110.847 \pm 6.556$\\ \cline{2-8} 
                                         & IsolFor & $0.039 \pm 0.000$ & $0.500 \pm 0.000$ & $0.039 \pm 0.000$ & $\bm{ 1.000 \pm 0.000 }$ & $0.076 \pm 0.000$ & $274.933 \pm 3.596$\\ \cline{2-8} 
                                         & OCRF & $0.996 \pm 0.000$ & $\bm{ 0.997 \pm 0.000 }$ & $0.958 \pm 0.000$ & $1.000 \pm 0.000$ & $ 0.977 \pm 0.000 $ & $160.931 \pm 196.420$\\ \cline{2-8} 
                                         & PNB & $0.354 \pm 0.227$ & $0.635 \pm 0.139$ & $0.486 \pm 0.419$ & $0.845 \pm 0.189$ & $0.454 \pm 0.252$ & $81.234 \pm 50.569$\\ \cline{2-8}
                             & GloVe CVDD & $0.989 \pm 0.000$ & $ 0.914 \pm 0.002$ & $ \bm{0.992 \pm 0.000}$ & $ 0.997 \pm 0.000$ & $ \bm{0.994 \pm 0.000}$ & $ 267.377 \pm 27.111$\\ \cline{2-8}
                             & BERT CVDD & $0.974 \pm 0.124$ & $ 0.906 \pm 0.259$ & $ 0.986 \pm 0.192$ & $ 0.974 \pm 0.175$ & $ 0.982 \pm 0.184$ & $ 307.439 \pm 37.084$\\ \cline{2-8}
                                         & CC & $\bm{ 0.997 \pm 0.008 }$ & $0.989 \pm 0.004$ & $0.970 \pm 0.017$ & $0.980 \pm 0.096$ & $0.973 \pm 0.001$ & $\bm{ 11.435 \pm 0.062 }$\\ \specialrule{.1em}{.05em}{.05em}  

\multirow{8}{*}{ MoviePlots } & Linear OCSVM & $0.635 \pm 0.035$ & $0.595 \pm 0.026$ & $0.489 \pm 0.054$ & $0.470 \pm 0.215$ & $0.440 \pm 0.106$ & $664.580 \pm 4.535$\\ \cline{2-8} 
                              & RBF OCSVM & $0.596 \pm 0.078$ & $0.589 \pm 0.022$ & $0.465 \pm 0.061$ & $0.565 \pm 0.276$ & $0.458 \pm 0.097$ & $718.690 \pm 5.346$\\ \cline{2-8} 
                              & Sigmoid OCSVM & $0.658 \pm 0.019$ & $0.570 \pm 0.031$ & $0.530 \pm 0.049$ & $0.294 \pm 0.184$ & $0.336 \pm 0.127$ & $699.067 \pm 11.559$\\ \cline{2-8} 
                              & Poly OCSVM & $0.617 \pm 0.084$ & $0.584 \pm 0.030$ & $0.523 \pm 0.098$ & $0.478 \pm 0.337$ & $0.402 \pm 0.151$ & $624.270 \pm 3.389$\\ \cline{2-8} 
                              & IsolFor & $0.339 \pm 0.000$ & $0.500 \pm 0.000$ & $0.339 \pm 0.000$ & $\bm{ 1.000 \pm 0.000 }$ & $0.507 \pm 0.000$ & $262.591 \pm 13.770$\\ \cline{2-8} 
                              & OCRF & $0.851 \pm 0.000$ & $0.895 \pm 0.000$ & $0.660 \pm 0.000$ & $0.922 \pm 0.003$ & $0.795 \pm 0.000$ & $161.427 \pm 197.625$\\ \cline{2-8} 
                              & PNB & $0.875 \pm 0.143$ & $0.910 \pm 0.068$ & $\bm{ 0.972 \pm 0.022 }$ & $0.885 \pm 0.156$ & $0.917 \pm 0.099$ & $281.049 \pm 22.044$\\ \cline{2-8}
                             & GloVe CVDD & $0.822 \pm 0.020$ & $ 0.801 \pm 0.020$ & $ 0.799 \pm 0.018$ & $ 0.928 \pm 0.020$ & $ 0.859 \pm 0.017$ & $ 274.335 \pm 56.185$\\ \cline{2-8}
                             & BERT CVDD & $0.817 \pm 0.020$ & $ 0.789 \pm 0.020$ & $ 0.776 \pm 0.018$ & $ 0.905 \pm 0.020$ & $ 0.843 \pm 0.017$ & $ 314.118 \pm 46.387$\\ \cline{2-8}
                              & CC & $\bm{ 0.953 \pm 0.004 }$ & $\bm{ 0.947 \pm 0.003 }$ & $0.934 \pm 0.011$ & $0.931 \pm 0.008$ & $\bm{ 0.931 \pm 0.005 }$ & $\bm{ 17.512 \pm 0.028 }$\\ \specialrule{.1em}{.05em}{.05em}  

\multirow{8}{*}{ Wikileaks } & Linear OCSVM & $0.999 \pm 0.008$ & $0.717 \pm 0.058$ & $1.000 \pm 0.000$ & $0.434 \pm 0.117$ & $0.596 \pm 0.117$ & $9.993 \pm 0.152$\\ \cline{2-8} 
                             & RBF OCSVM & $0.999 \pm 0.005$ & $0.708 \pm 0.037$ & $1.000 \pm 0.000$ & $0.416 \pm 0.074$ & $0.584 \pm 0.072$ & $11.337 \pm 0.272$\\ \cline{2-8} 
                             & Sigmoid OCSVM & $0.999 \pm 0.003$ & $0.745 \pm 0.024$ & $1.000 \pm 0.000$ & $0.491 \pm 0.040$ & $0.658 \pm 0.037$ & $10.628 \pm 0.356$\\ \cline{2-8} 
                             & Poly OCSVM & $0.999 \pm 0.002$ & $0.680 \pm 0.019$ & $1.000 \pm 0.000$ & $0.361 \pm 0.039$ & $0.529 \pm 0.041$ & $\bm{ 9.783 \pm 0.167 }$\\ \cline{2-8} 
                             & IsolFor & $0.999 \pm 0.008$ & $0.833 \pm 0.058$ & $1.000 \pm 0.000$ & $0.667 \pm 0.117$ & $0.794 \pm 0.081$ & $287.498 \pm 1.478$\\ \cline{2-8} 
                             & OCRF & $0.999 \pm 0.000$ & $1.000 \pm 0.000$ & $0.999 \pm 0.000$ & $1.000 \pm 0.000$ & $0.999 \pm 0.000$ & $164.122 \pm 201.317$\\ \cline{2-8} 
                             & PNB & $0.087 \pm 0.244$ & $0.541 \pm 0.122$ & $0.070 \pm 0.248$ & $1.000 \pm 0.000$ & $0.074 \pm 0.247$ & $78.363 \pm 49.268$\\ \cline{2-8}
                             & GloVe CVDD & $1.000 \pm 0.000$ & $ 1.000 \pm 0.000$ & $ 1.000 \pm 0.000$ & $ 1.000 \pm 0.000$ & $ 1.000 \pm 0.000$ & $ 253.909 \pm 28.636$\\ \cline{2-8}
                             & BERT CVDD & $1.000 \pm 0.000$ & $ 1.000 \pm 0.000$ & $ 1.000 \pm 0.000$ & $ 1.000 \pm 0.000$ & $ 1.000 \pm 0.000$ & $ 298.142 \pm 38.494$\\ \cline{2-8}
                             & CC & $\bm{ 1.000 \pm 0.000 }$ & $\bm{ 1.000 \pm 0.000 }$ & $\bm{ 1.000 \pm 0.000 }$ & $\bm{ 1.000 \pm 0.000 }$ & $\bm{ 1.000 \pm 0.000 }$ & $11.163 \pm 0.014$
                             \\ \specialrule{.1em}{.05em}{.05em} 

\end{tabular}%
}
\vspace{-4mm}
\end{table*}

\subsection{Evaluation Setup}
When a given dataset is being evaluated as the positive class, the rest of the datasets are combined and treated as the negative class. Since our training set does not require any data from the negative class, we split each class via a 50\%-50\% split between our validation and test sets. Our positive class is split using a 70\%-15\%-15\% split between our training set, our validation set, and our test set. Resplitting our train and test sets each run, we compile evaluation metrics accuracy, balanced accuracy,  precision, recall, F1 score, and time 20 times per dataset. We report mean and standard deviation values.

In order to be able to compare compute times, all models will be run on the same free instance of Google Colaboratory \cite{colab}. Our evaluation instance had a single core running at 2.00GHz, and had access to 13 Gb of RAM.

Finally, we discuss the various VSM models used, comparing the baseline VSMs with NE-TF.

\section{Results}
Performance metrics can be found in Table 1.

\subsection{Predictive Power}
CC outperforms baseline models in most scenarios, being the only model with mean accuracies consistently above 95\%, balanced accuracies above 94\%, and precision, recall, and F1 scores above 93\%. PNB had the largest variability out of all algorithms both on a dataset level as well as on a per run level, showcasing how dependent it is on the exact distribution of words that exist within the unlabeled set. OCRF is one of the best performing baseline models, while IsolFor performed the worst, clearly showing that the splitting algorithm used to determine tree structure is crucial for topic determination with ensemble models. The Linear OCSVM outperformed OCSVM alternatives.

The performance delta between BERT and GloVe does not justify the additional computation costs involved with using a BERT encoding for our problem space. Both neural NLP models are consistently outperformed by CC across datasets for one class topic determination. While CVDD and other neural NLP algorithms that use embeddings have use cases in one-class topic determination where they work well, they perform worse when the positive class is highly manifold in nature as is the case for the Jobs and MoviePlots datasets.

\subsection{Computation Efficiency}
Where CC truly shines is in computational efficiency, showcased in the scenarios with high text complexity. Since we compare each evaluation vector to the max and min vectors, CC has a worse case runtime efficiency of O(dn), where d is the vector dimension number and n is the number of vectors to be evaluated. In practice however, the efficiency is much greater, as we can short-circuit computation as soon as we find a discrepancy; this is a benefit that none of the baselines have. When we compare this to ensembles with a runtime of O(d*nlog(n)), kernel OCSVMs with a runtime of O($n_{support}$*dn) where $n_{support}$ is the number of support vectors, PNB which has a runtime of O(dn + 4d) due to performing training and evaluation at the same time, and neural NLP solutions having a forward pass complexity of $O(n^4)$ \cite{fredenslund2018computational}, the efficiency of CC is clear.

Linear OCSVM has the highest computation efficiency out of the baselines, with a similar worse case runtime efficiency as CC of O(dn). However for each vector at each dimension, Linear OCSVM performs two operations compared to only one, a multiplication as well as an addition. Additionally, Linear OCSVM has no short-circuit capability, so it will always take the maximal amount of time to compute. This can be seen in our results, where CC outperforms Linear OCSVM in computation time especially on the more complex datasets like MedicalTranscriptions and MoviePlots where the time differences are stark.

\subsection{VSM Comparison}
We identified that the encoding and embedding process is the foremost reason behind the long computation times both versions of CVDD has. This is the reason behind the development of NE-TF; being a bag-of-words VSM it boasts great speed in creating its vector representations. Additionally, bag-of-words VSM models like NE-TF also provides benefits in terms of memory footprint; for our datasets, SpacyEncoding requires 154.7MB, BertTokenizer requires 157.1MB, while NE-TF requires only 18.3MB leading to a roughly 9 times smaller footprint. 

When we compare to alternative bag-of-words VSM models NE-TF has a comparable memory footprint but is faster to compute; the statistical significance weighting mitigates the need for stop word pruning, further improving performance.

\section{Conclusion}
We show that Conical Classification is a computationally efficient method of one-class topic classification that aims to identify whether a vector is within the conical span of the training corpus for a given topic. When combined with Normal Exclusion, Conical Classification showcases the optimal combination of predictive power, consistently great results, and fast computation times. 

\nocite{scikit-learn}
\bibliography{anthology,custom}

\begin{thebibliography}{64}
\expandafter\ifx\csname natexlab\endcsname\relax\def\natexlab#1{#1}\fi

\bibitem[{Agarap and Grafilon(2018)}]{agarap2018statistical}
Abien~Fred Agarap and Paul Grafilon. 2018.
\newblock Statistical analysis on e-commerce reviews, with sentiment
  classification using bidirectional recurrent neural network (rnn).
\newblock \emph{arXiv preprint arXiv:1805.03687}.

\bibitem[{Ahmed et~al.(2017)Ahmed, Traore, and Saad}]{ahmed2017detection}
Hadeer Ahmed, Issa Traore, and Sherif Saad. 2017.
\newblock Detection of online fake news using n-gram analysis and machine
  learning techniques.
\newblock In \emph{International conference on intelligent, secure, and
  dependable systems in distributed and cloud environments}, pages 127--138.
  Springer.

\bibitem[{Ahmed et~al.(2018)Ahmed, Traore, and Saad}]{ahmed2018detecting}
Hadeer Ahmed, Issa Traore, and Sherif Saad. 2018.
\newblock Detecting opinion spams and fake news using text classification.
\newblock \emph{Security and Privacy}, 1(1):e9.

\bibitem[{Alghamdi et~al.(2019)Alghamdi, Alharby
  et~al.}]{alghamdi2019intelligent}
Bandar Alghamdi, Fahad Alharby, et~al. 2019.
\newblock An intelligent model for online recruitment fraud detection.
\newblock \emph{Journal of Information Security}, 10(03):155.

\bibitem[{Anderka et~al.(2011)Anderka, Stein, and Lipka}]{anderka2011detection}
Maik Anderka, Benno Stein, and Nedim Lipka. 2011.
\newblock Detection of text quality flaws as a one-class classification
  problem.
\newblock In \emph{Proceedings of the 20th ACM international conference on
  Information and knowledge management}, pages 2313--2316.

\bibitem[{Balachander and Moh(2018)}]{balachander2018ontology}
Yeshwanth Balachander and Teng-Sheng Moh. 2018.
\newblock Ontology based similarity for information technology skills.
\newblock In \emph{2018 IEEE/ACM International Conference on Advances in Social
  Networks Analysis and Mining (ASONAM)}, pages 302--305. IEEE.

\bibitem[{Beattie and Richards(1994)}]{beattie1994separation}
John~H Beattie and Mark~P Richards. 1994.
\newblock Separation of metallothionein isoforms by micellar electrokinetic
  capillary chromatography.
\newblock \emph{Journal of Chromatography A}, 664(1):129--134.

\bibitem[{Brants and Franz(2006)}]{brantsgooglewordset}
Thorsten Brants and Alex Franz. 2006.
\newblock 1t 5-gram version 1.0.
\newblock \emph{Linguistic Data Consortium}.

\bibitem[{Cascaro et~al.(2019)Cascaro, Gerardo, and
  Medina}]{cascaro2019aggregating}
Rhodessa~J Cascaro, Bobby~D Gerardo, and Ruji~P Medina. 2019.
\newblock Aggregating filter feature selection methods to enhance multiclass
  text classification.
\newblock In \emph{Proceedings of the 2019 7th International Conference on
  Information Technology: IoT and Smart City}, pages 80--84.

\bibitem[{Chattopadhyay et~al.(2018)Chattopadhyay, Wang, and
  Tan}]{chattopadhyay2018scenario}
Pratik Chattopadhyay, Lipo Wang, and Yap-Peng Tan. 2018.
\newblock Scenario-based insider threat detection from cyber activities.
\newblock \emph{IEEE Transactions on Computational Social Systems},
  5(3):660--675.

\bibitem[{Davis(1954)}]{davis1954theory}
Chandler Davis. 1954.
\newblock Theory of positive linear dependence.
\newblock \emph{American Journal of Mathematics}, 76(4):733--746.

\bibitem[{Denis et~al.(2003)Denis, Laurent, Gilleron, and
  Tommasi}]{denis2003text}
Francois Denis, Anne Laurent, R{\'e}mi Gilleron, and Marc Tommasi. 2003.
\newblock Text classification and co-training from positive and unlabeled
  examples.
\newblock In \emph{Proceedings of the ICML 2003 workshop: the continuum from
  labeled to unlabeled data}, pages 80--87.

\bibitem[{Devlin et~al.(2018)Devlin, Chang, Lee, and
  Toutanova}]{devlin2018bert}
Jacob Devlin, Ming-Wei Chang, Kenton Lee, and Kristina Toutanova. 2018.
\newblock Bert: Pre-training of deep bidirectional transformers for language
  understanding.
\newblock \emph{arXiv preprint arXiv:1810.04805}.

\bibitem[{Domeniconi et~al.(2015)Domeniconi, Moro, Pasolini, and
  Sartori}]{domeniconi2015comparison}
Giacomo Domeniconi, Gianluca Moro, Roberto Pasolini, and Claudio Sartori. 2015.
\newblock A comparison of term weighting schemes for text classification and
  sentiment analysis with a supervised variant of tf. idf.
\newblock In \emph{International Conference on Data Management Technologies and
  Applications}, pages 39--58. Springer.

\bibitem[{Forman(2008)}]{forman2008bns}
George Forman. 2008.
\newblock Bns feature scaling: an improved representation over tf-idf for svm
  text classification.
\newblock In \emph{Proceedings of the 17th ACM conference on Information and
  knowledge management}, pages 263--270.

\bibitem[{Forman et~al.(2003)}]{forman2003extensive}
George Forman et~al. 2003.
\newblock An extensive empirical study of feature selection metrics for text
  classification.
\newblock \emph{J. Mach. Learn. Res.}, 3(Mar):1289--1305.

\bibitem[{Fredenslund(2018)}]{fredenslund2018computational}
K~Fredenslund. 2018.
\newblock Computational complexity of neural networks.

\bibitem[{Glasser and Lindauer(2013)}]{glasser2013bridging}
Joshua Glasser and Brian Lindauer. 2013.
\newblock Bridging the gap: A pragmatic approach to generating insider threat
  data.
\newblock In \emph{2013 IEEE Security and Privacy Workshops}, pages 98--104.
  IEEE.

\bibitem[{Goix et~al.(2017)Goix, Drougard, Brault, and Chiapino}]{goix2017one}
Nicolas Goix, Nicolas Drougard, Romain Brault, and Mael Chiapino. 2017.
\newblock One class splitting criteria for random forests.
\newblock In \emph{Asian Conference on Machine Learning}, pages 343--358. PMLR.

\bibitem[{Google(2019)}]{colab}
Google. 2019.
\newblock Google colaboratory.

\bibitem[{Hakim et~al.(2014)Hakim, Erwin, Eng, Galinium, and
  Muliady}]{hakim2014automated}
Ari~Aulia Hakim, Alva Erwin, Kho~I Eng, Maulahikmah Galinium, and Wahyu
  Muliady. 2014.
\newblock Automated document classification for news article in bahasa
  indonesia based on term frequency inverse document frequency (tf-idf)
  approach.
\newblock In \emph{2014 6th international conference on information technology
  and electrical engineering (ICITEE)}, pages 1--4. IEEE.

\bibitem[{Hempstalk et~al.(2008)Hempstalk, Frank, and
  Witten}]{hempstalk2008one}
Kathryn Hempstalk, Eibe Frank, and Ian~H Witten. 2008.
\newblock One-class classification by combining density and class probability
  estimation.
\newblock In \emph{Joint European Conference on Machine Learning and Knowledge
  Discovery in Databases}, pages 505--519. Springer.

\bibitem[{Hu et~al.(2021)Hu, Feng, Kamigaito, Takamura, and
  Okumura}]{hu-etal-2021-one}
Chenlong Hu, Yukun Feng, Hidetaka Kamigaito, Hiroya Takamura, and Manabu
  Okumura. 2021.
\newblock \href {https://aclanthology.org/2021.eacl-main.296} {One-class text
  classification with multi-modal deep support vector data description}.
\newblock In \emph{Proceedings of the 16th Conference of the European Chapter
  of the Association for Computational Linguistics: Main Volume}, pages
  3378--3390, Online. Association for Computational Linguistics.

\bibitem[{Joffe et~al.(2015)Joffe, Pettigrew, Herskovic, Bearden, and
  Bernstam}]{joffe2015expert}
Erel Joffe, Emily~J Pettigrew, Jorge~R Herskovic, Charles~F Bearden, and
  Elmer~V Bernstam. 2015.
\newblock Expert guided natural language processing using one-class
  classification.
\newblock \emph{Journal of the American Medical Informatics Association},
  22(5):962--966.

\bibitem[{Jones(1972)}]{jones1972statistical}
Karen~Sparck Jones. 1972.
\newblock A statistical interpretation of term specificity and its application
  in retrieval.
\newblock \emph{Journal of documentation}.

\bibitem[{Kim et~al.(2019{\natexlab{a}})Kim, Seo, Cho, and Kang}]{kim2019multi}
Donghwa Kim, Deokseong Seo, Suhyoun Cho, and Pilsung Kang. 2019{\natexlab{a}}.
\newblock Multi-co-training for document classification using various document
  representations: Tf--idf, lda, and doc2vec.
\newblock \emph{Information Sciences}, 477:15--29.

\bibitem[{Kim et~al.(2019{\natexlab{b}})Kim, Kim, and Kim}]{kim2019fraud}
Jeongrae Kim, Han-Joon Kim, and Hyoungrae Kim. 2019{\natexlab{b}}.
\newblock Fraud detection for job placement using hierarchical clusters-based
  deep neural networks.
\newblock \emph{Applied Intelligence}, 49(8):2842--2861.

\bibitem[{Kim and Gil(2019)}]{kim2019research}
Sang-Woon Kim and Joon-Min Gil. 2019.
\newblock Research paper classification systems based on tf-idf and lda
  schemes.
\newblock \emph{Human-centric Computing and Information Sciences}, 9(1):1--21.

\bibitem[{Kousta and Bellet()}]{koustalocal}
Theodora Kousta and Clement~S Bellet.
\newblock Local interpretable model-agnostic explanations for long short-term
  memory network used for classification of amazon customer reviews.

\bibitem[{Le et~al.(2018)Le, Khanchi, Zincir-Heywood, and
  Heywood}]{le2018benchmarking}
Duc~C Le, Sara Khanchi, A~Nur Zincir-Heywood, and Malcolm~I Heywood. 2018.
\newblock Benchmarking evolutionary computation approaches to insider threat
  detection.
\newblock In \emph{Proceedings of the Genetic and Evolutionary Computation
  Conference}, pages 1286--1293.

\bibitem[{Le and Zincir-Heywood(2019)}]{le2019machine}
Duc~C Le and A~Nur Zincir-Heywood. 2019.
\newblock Machine learning based insider threat modelling and detection.
\newblock In \emph{2019 IFIP/IEEE Symposium on Integrated Network and Service
  Management (IM)}, pages 1--6. IEEE.

\bibitem[{Le et~al.(2021)Le, Noumeir, Rambaud, Sans, and
  Jouvet}]{le2021machine}
Thanh-Dung Le, Rita Noumeir, Jerome Rambaud, Guillaume Sans, and Philippe
  Jouvet. 2021.
\newblock Machine learning based on natural language processing to detect
  cardiac failure in clinical narratives.
\newblock \emph{arXiv preprint arXiv:2104.03934}.

\bibitem[{Lee et~al.(2011)Lee, Palsetia, Narayanan, Patwary, Agrawal, and
  Choudhary}]{lee2011twitter}
Kathy Lee, Diana Palsetia, Ramanathan Narayanan, Md~Mostofa~Ali Patwary, Ankit
  Agrawal, and Alok Choudhary. 2011.
\newblock Twitter trending topic classification.
\newblock In \emph{2011 IEEE 11th International Conference on Data Mining
  Workshops}, pages 251--258. IEEE.

\bibitem[{Lin(2020)}]{lin2020sentiment}
Xiaoxin Lin. 2020.
\newblock Sentiment analysis of e-commerce customer reviews based on natural
  language processing.
\newblock In \emph{Proceedings of the 2020 2nd International Conference on Big
  Data and Artificial Intelligence}, pages 32--36.

\bibitem[{Liu et~al.(2018)Liu, Sheng, Wei, and Yang}]{liu2018research}
Cai-zhi Liu, Yan-xiu Sheng, Zhi-qiang Wei, and Yong-Quan Yang. 2018.
\newblock Research of text classification based on improved tf-idf algorithm.
\newblock In \emph{2018 IEEE International Conference of Intelligent Robotic
  and Control Engineering (IRCE)}, pages 218--222. IEEE.

\bibitem[{Liu et~al.(2008)Liu, Ting, and Zhou}]{liu2008isolation}
Fei~Tony Liu, Kai~Ming Ting, and Zhi-Hua Zhou. 2008.
\newblock Isolation forest.
\newblock In \emph{2008 eighth ieee international conference on data mining},
  pages 413--422. IEEE.

\bibitem[{Mahbub and Pardede(2018)}]{mahbub2018using}
Syed Mahbub and Eric Pardede. 2018.
\newblock Using contextual features for online recruitment fraud detection.

\bibitem[{Manevitz and Yousef(2001)}]{manevitz2001one}
Larry~M Manevitz and Malik Yousef. 2001.
\newblock One-class svms for document classification.
\newblock \emph{Journal of machine Learning research}, 2(Dec):139--154.

\bibitem[{Martineau and Finin(2009)}]{martineau2009delta}
Justin Martineau and Tim Finin. 2009.
\newblock Delta tfidf: An improved feature space for sentiment analysis.
\newblock In \emph{Proceedings of the International AAAI Conference on Web and
  Social Media}, volume~3.

\bibitem[{Meng et~al.(2018)Meng, Lou, Fu, and Tian}]{meng2018deep}
Fanzhi Meng, Fang Lou, Yunsheng Fu, and Zhihong Tian. 2018.
\newblock Deep learning based attribute classification insider threat detection
  for data security.
\newblock In \emph{2018 IEEE Third International Conference on Data Science in
  Cyberspace (DSC)}, pages 576--581. IEEE.

\bibitem[{Moramarco et~al.(2021)Moramarco, Juric, Savkov, and
  Reiter}]{moramarco2021towards}
Francesco Moramarco, Damir Juric, Aleksandar Savkov, and Ehud Reiter. 2021.
\newblock Towards objectively evaluating the quality of generated medical
  summaries.
\newblock \emph{arXiv preprint arXiv:2104.04412}.

\bibitem[{MTSamples()}]{medtransdataset}
MTSamples.
\newblock Collection of transcribed medical transcription sample reports and
  examples.

\bibitem[{Oxford(2011)}]{oecstats}
Oxford. 2011.
\newblock The oec: Facts about the language.
\newblock \emph{Oxford English Dictionary}.

\bibitem[{Pedregosa et~al.(2011)Pedregosa, Varoquaux, Gramfort, Michel,
  Thirion, Grisel, Blondel, Prettenhofer, Weiss, Dubourg, Vanderplas, Passos,
  Cournapeau, Brucher, Perrot, and Duchesnay}]{scikit-learn}
F.~Pedregosa, G.~Varoquaux, A.~Gramfort, V.~Michel, B.~Thirion, O.~Grisel,
  M.~Blondel, P.~Prettenhofer, R.~Weiss, V.~Dubourg, J.~Vanderplas, A.~Passos,
  D.~Cournapeau, M.~Brucher, M.~Perrot, and E.~Duchesnay. 2011.
\newblock Scikit-learn: Machine learning in {P}ython.
\newblock \emph{Journal of Machine Learning Research}, 12:2825--2830.

\bibitem[{Pennington et~al.(2014)Pennington, Socher, and
  Manning}]{pennington2014glove}
Jeffrey Pennington, Richard Socher, and Christopher~D Manning. 2014.
\newblock Glove: Global vectors for word representation.
\newblock In \emph{Proceedings of the 2014 conference on empirical methods in
  natural language processing (EMNLP)}, pages 1532--1543.

\bibitem[{PromptCloud(2017)}]{monsterjobs17}
PromptCloud. 2017.
\newblock Us jobs kaggle dataset.

\bibitem[{Rao and Gudivada(2018)}]{rao2018computational}
CR~Rao and Venkat~N Gudivada. 2018.
\newblock \emph{Computational analysis and understanding of natural languages:
  Principles, methods and applications}.
\newblock Elsevier.

\bibitem[{Reddy et~al.(2018)Reddy, Mamatha, and Balaram}]{reddy2018analysis}
M~Niharika Reddy, T~Mamatha, and A~Balaram. 2018.
\newblock Analysis of e-recruitment systems and detecting e-recruitment fraud.
\newblock In \emph{International Conference on Communications and Cyber
  Physical Engineering 2018}, pages 411--417. Springer.

\bibitem[{Ruff et~al.(2019)Ruff, Zemlyanskiy, Vandermeulen, Schnake, and
  Kloft}]{ruff2019self}
Lukas Ruff, Yury Zemlyanskiy, Robert Vandermeulen, Thomas Schnake, and Marius
  Kloft. 2019.
\newblock Self-attentive, multi-context one-class classification for
  unsupervised anomaly detection on text.
\newblock In \emph{Proceedings of the 57th Annual Meeting of the Association
  for Computational Linguistics}, pages 4061--4071.

\bibitem[{Sajjanhar et~al.(2019)Sajjanhar, Xiang et~al.}]{sajjanhar2019image}
Atul Sajjanhar, Yong Xiang, et~al. 2019.
\newblock Image-based feature representation for insider threat classification.
\newblock \emph{arXiv preprint arXiv:1911.05879}.

\bibitem[{Seo(2007)}]{seo2007application}
Kwang-Kyu Seo. 2007.
\newblock An application of one-class support vector machines in content-based
  image retrieval.
\newblock \emph{Expert Systems with Applications}, 33(2):491--498.

\bibitem[{Sitikhu et~al.(2019)Sitikhu, Pahi, Thapa, and
  Shakya}]{sitikhu2019comparison}
Pinky Sitikhu, Kritish Pahi, Pujan Thapa, and Subarna Shakya. 2019.
\newblock A comparison of semantic similarity methods for maximum human
  interpretability.
\newblock In \emph{2019 artificial intelligence for transforming business and
  society (AITB)}, volume~1, pages 1--4. IEEE.

\bibitem[{Stappen(2020)}]{positivespanchart}
Frank Stappen. 2020.
\newblock Motion and manipulation lecture series.
\newblock \emph{Utrecht University}.

\bibitem[{Sun et~al.(2019)Sun, Jiang, and Dai}]{sun2019sentiment}
Hao Sun, Tao Jiang, and Yugang Dai. 2019.
\newblock Sentiment analysis of commodity reviews based on multilayer lstm
  network.
\newblock In \emph{Proceedings of the International Conference on Artificial
  Intelligence, Information Processing and Cloud Computing}, pages 1--5.

\bibitem[{Tatman(2017)}]{tatmankaggleword}
Rachael Tatman. 2017.
\newblock $\frac{1}{3}$ million most frequent english words on the web.
\newblock \emph{Kaggle}.

\bibitem[{Trstenjak et~al.(2014)Trstenjak, Mikac, and Donko}]{trstenjak2014knn}
Bruno Trstenjak, Sasa Mikac, and Dzenana Donko. 2014.
\newblock Knn with tf-idf based framework for text categorization.
\newblock \emph{Procedia Engineering}, 69:1356--1364.

\bibitem[{Tuor et~al.(2017)Tuor, Kaplan, Hutchinson, Nichols, and
  Robinson}]{tuor2017deep}
Aaron Tuor, Samuel Kaplan, Brian Hutchinson, Nicole Nichols, and Sean Robinson.
  2017.
\newblock Deep learning for unsupervised insider threat detection in structured
  cybersecurity data streams.
\newblock \emph{arXiv preprint arXiv:1710.00811}.

\bibitem[{Vidros et~al.(2017)Vidros, Kolias, Kambourakis, and
  Akoglu}]{vidros2017automatic}
Sokratis Vidros, Constantinos Kolias, Georgios Kambourakis, and Leman Akoglu.
  2017.
\newblock Automatic detection of online recruitment frauds: Characteristics,
  methods, and a public dataset.
\newblock \emph{Future Internet}, 9(1):6.

\bibitem[{Wang and Manning(2012)}]{wang2012baselines}
Sida~I Wang and Christopher~D Manning. 2012.
\newblock Baselines and bigrams: Simple, good sentiment and topic
  classification.
\newblock In \emph{Proceedings of the 50th Annual Meeting of the Association
  for Computational Linguistics (Volume 2: Short Papers)}, pages 90--94.

\bibitem[{Wei et~al.(2021)Wei, Chow, and Yiu}]{wei2021insider}
Yichen Wei, Kam-Pui~P Chow, and Siu-Ming Yiu. 2021.
\newblock Insider threat prediction based on unsupervised anomaly detection
  scheme for proactive forensic investigation.
\newblock \emph{Digital Investigation}.

\bibitem[{Xiao and Tong(2021)}]{Xiao_2021}
Shuwei Xiao and Weiqin Tong. 2021.
\newblock Prediction of user consumption behavior data based on the combined
  model of {TF}-{IDF} and logistic regression.
\newblock \emph{Journal of Physics: Conference Series}, 1757(1):012089.

\bibitem[{Zhang et~al.(2011)Zhang, Yoshida, and Tang}]{zhang2011comparative}
Wen Zhang, Taketoshi Yoshida, and Xijin Tang. 2011.
\newblock A comparative study of tf* idf, lsi and multi-words for text
  classification.
\newblock \emph{Expert Systems with Applications}, 38(3):2758--2765.

\bibitem[{Zhuang and Dai(2006)}]{zhuang2006parameter}
Ling Zhuang and Honghua Dai. 2006.
\newblock Parameter estimation of one-class svm on imbalance text
  classification.
\newblock In \emph{Conference of the Canadian Society for Computational Studies
  of Intelligence}, pages 538--549. Springer.

\bibitem[{Zuccon et~al.(2014)Zuccon, Kotzur, Nguyen, and
  Bergheim}]{zuccon2014identification}
Guido Zuccon, Daniel Kotzur, Anthony Nguyen, and Anton Bergheim. 2014.
\newblock De-identification of health records using anonym: Effectiveness and
  robustness across datasets.
\newblock \emph{Artificial intelligence in medicine}, 61(3):145--151.

\end{thebibliography}
\bibliographystyle{acl_natbib}
\end{document}